\documentclass[conference]{IEEEtran}
\IEEEoverridecommandlockouts
\usepackage{cite}
\usepackage{amsmath,amssymb,amsfonts}
\usepackage{algorithmic}
\usepackage{graphicx}
\usepackage{textcomp}
\usepackage{xcolor}
\usepackage{float}
\usepackage{multirow}
\usepackage{booktabs}
\def\BibTeX{{\rm B\kern-.05em{\sc i\kern-.025em b}\kern-.08em
    T\kern-.1667em\lower.7ex\hbox{E}\kern-.125emX}}
\begin{document}

\title{CasDyF-Net: Image Dehazing via Cascaded Dynamic Filters\\

}

\author{\IEEEauthorblockN{1\textsuperscript{st} Yinglong Wang}
\IEEEauthorblockA{\textit{School of Computing and Artificial Intelligence} \\
\textit{Southwest Jiaotong University}\\
Chengdu, China \\
wangyinglong2023@gmail.com}
\and
\IEEEauthorblockN{2\textsuperscript{nd} Bin He\textsuperscript{*}}
\IEEEauthorblockA{\textit{School of Computing and Artificial Intelligence} \\
\textit{Southwest Jiaotong University}\\
Chengdu, China \\
bhe@home.swjtu.edu.cn}
}

\maketitle

\begin{abstract}
    Image dehazing aims to restore image clarity and visual quality by reducing atmospheric scattering and absorption effects. While deep learning has made significant strides in this area, more and more methods are constrained by network depth. Consequently, lots of approaches have adopted parallel branching strategies. however, they often prioritize aspects such as resolution, receptive field, or frequency domain segmentation without dynamically partitioning branches based on the distribution of input features. Inspired by dynamic filtering, we propose using cascaded dynamic filters to create a multi-branch network by dynamically generating filter kernels based on feature map distribution. To better handle branch features, we propose a residual multiscale block (RMB), combining different receptive fields. Furthermore, We also introduce a dynamic convolution-based local fusion method to merge features from adjacent branches. Experiments on RESIDE, Haze4K, and O-Haze datasets validate our method’s effectiveness, with our model achieving a PSNR of 43.21dB on the RESIDE-Indoor dataset. The code is available at https://github.com/dauing/CasDyF-Net.
\end{abstract}


\section{Introduction}

Image dehazing is vital in computer vision, addressing atmospheric haziness caused by particles like water vapor, smoke, and dust\cite{b7}. This haziness degrades image quality, complicating tasks such as object detection, semantic segmentation, and autonomous driving. Traditional methods\cite{b2}\cite{b3}\cite{b4} rely on prior knowledge, which limits their generalizability across varying lighting and haze densities, and they often suffer from computational inaccuracies\cite{b8}.

The rise of deep learning has revolutionized image dehazing. Techniques based on Convolutional Neural Networks (CNNs) and Transformers\cite{b33} outperform traditional methods in feature extraction, end-to-end learning, and generalization. CNN-based methods excel at capturing local features with tools like large kernels\cite{b25}, dilated convolutions\cite{b26}, and attention mechanisms\cite{b27}, which enhance dehazing robustness across diverse conditions. Transformers leverage self-attention to encapsulate long-range dependencies, aiding in the understanding of global image structures\cite{b10}.

\begin{figure}[t]
\centerline{\includegraphics[width=0.5\textwidth]{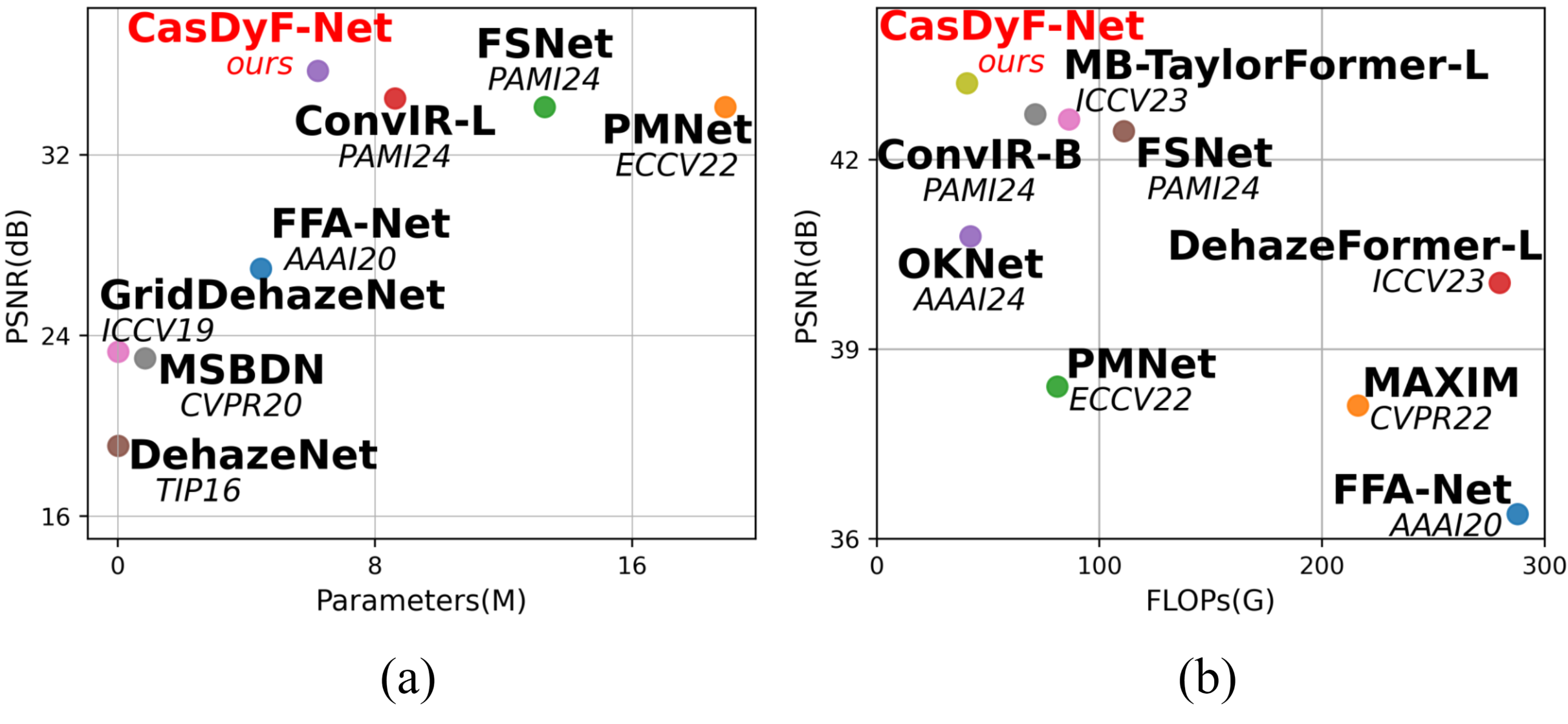}}
\caption{(a) Parameters vs. PSNR on the Haze4K dataset. (b) GFLOPs vs. PSNR on the SOTS-Indoor dataset. Our model achieves excellent performance with low computational overhead.}
\label{fig1}
\end{figure}

However, deeper CNNs face diminishing returns due to issues like vanishing gradients and accuracy degradation\cite{b13}. Multi-branch parallel networks offer a solution by allowing different branches to learn distinct features, which are then integrated to enhance representation capacity and reduce information loss\cite{b14}\cite{b51}\cite{b55}. Yet, existing methods for branch creation have limitations, such as losing high-frequency details in resolution-based approaches\cite{b51} or failing to adapt to varying image content in frequency-based methods\cite{b55}\cite{b56}.

Inspired by dynamic filter kernels\cite{b57}, we propose a multi-branch structure based on cascaded dynamic filtering. Each filter isolates specific frequency bands, and dynamically adjusts to extract richer features at different levels. This method overcomes the limitations of traditional branch creation, effectively handling complex hazy scenarios by capturing diverse frequency features.

We further introduced a Residual Multiscale Block (RMB) to refine features from multiple branches, preserving texture details and global features across varying receptive fields. By incorporating varying dilation rates, we enhance multi-scale information utilization during dehazing.

For efficient feature integration, we devised a local fusion module using 1×1 dynamic convolution to merge adjacent frequency bands. This strategy improves continuity and preserves band characteristics, while a subsequent parallel attention mechanism ensures effective global and local information fusion. This progressive approach ensures that the final dehazed images maintain both clarity and naturalness.

\begin{figure*}[t]
  \centering
  \includegraphics[width=\textwidth]{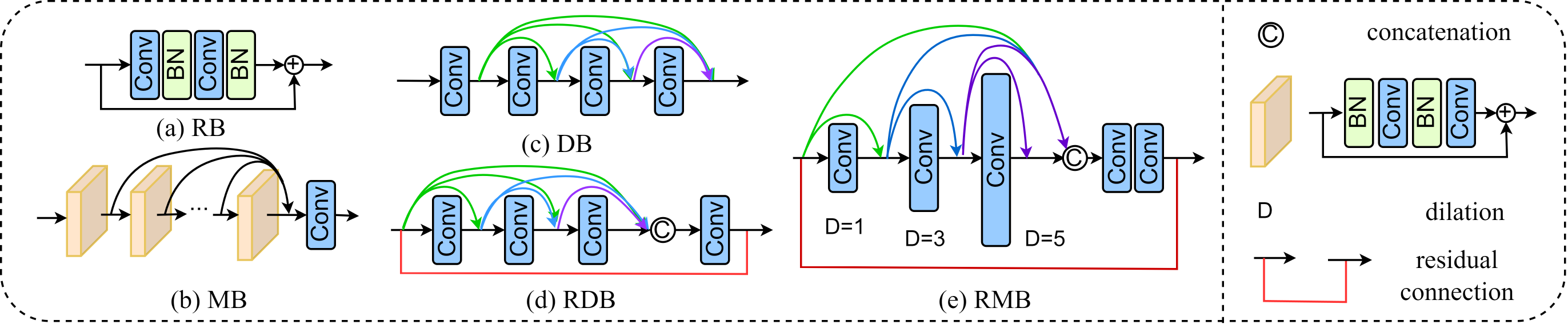} 
  \caption{Comparison of Several Convolutional Blocks in Convolutional Neural Networks: (a) residual block (RB) in SRResNet \cite{b28}, (b) memory block (MB) in MemNet \cite{b29}, (c) dense block (DB) in SRDenseNet \cite{b30}, (d) residual dense block (RDB) in RDN \cite{b31}, (e) proposed residual multiscale block (RMB).}
  \label{rmb}
\end{figure*}

Our model, tested on multiple datasets, significantly outperformed recent state-of-the-art models. For instance, compared to CNN-based FSNet, our model uses only 46.8\% of its parameters and achieves a 0.76dB improvement on the ITS dataset, while outperforming the Transformer-based MB-TaylorFormer with 83.9\% of its parameters and a 0.57dB gain. Key contributions include:

\begin{itemize}
\item A new dehazing architecture leveraging dynamic filter kernels for creating adaptable feature branches.
\item A Residual Multiscale Block (RMB) that enhances receptive fields and retains multi-scale information.
\item A local fusion method using dynamic convolution for improved performance with minimal computational cost.
\item A progressive fusion strategy combining local and global attention mechanisms for superior dehazing outcomes.
\item Comprehensive testing on public datasets, demonstrating the efficacy of our approach.
\end{itemize}

\section{Related Works}

\subsection{Image Dehazing}

Under adverse weather conditions like fog and haze, images often appear blurry, complicating visual tasks. Traditional image dehazing algorithms, usually rely on prior knowledge such as the Dark Channel Prior (DCP)\cite{b5}, attempt to restore clarity by estimating model parameters. However, these methods typically rely on manually designed features, which can limit their robustness\cite{b57}.

With the advent of deep learning, CNN-based approaches have significantly advanced image dehazing. Ren et al.\cite{b12} pioneered the use of CNNs for dehazing by employing multi-scale models to estimate the transmission map. Zhang et al.\cite{b16} later introduced end-to-end dehazing networks, incorporating advanced techniques such as residual learning\cite{b17}, attention mechanisms\cite{b22}\cite{b23}, and U-Net architectures\cite{b18}. Despite the effectiveness of large convolution kernels in capturing richer features\cite{b25}\cite{b27}, their computational overhead remains a challenge. To mitigate this, some studies have proposed approximating large kernels with standard and sparse convolutions\cite{b26}. However, CNNs often struggle with global haze characteristics due to their limited capacity to handle long-range dependencies.

Transformers\cite{b33}, leveraging self-attention, can capture these long-range dependencies, making them well-suited for global structure understanding in hazy images\cite{b10}. Recent studies combining Transformers with dehazing models have achieved promising results\cite{b34}\cite{b35}. However, the computational complexity of self-attention has driven efforts to develop approximation methods for simplifying Transformers\cite{b40}.

\subsection{Multi-Branch Networks}

Initially, increasing CNN depth was a primary strategy for enhancing model performance. However, deeper networks often introduced issues such as overfitting and gradient problems\cite{b13}. As a result, many researches shifted towards multi-branch networks, which utilize parallel processing paths to capture richer features at various levels\cite{b14}\cite{b51}\cite{b55}.

Despite their robustness, conventional multi-branch strategies, like those based on image resolution\cite{b57} or receptive fields\cite{b14}, have limitations. For instance, they may neglect semantic information or lead to redundant features. More recent approaches involve frequency-based methods, such as dynamic filters that dynamically separate frequency bands in images, allowing the model to better adapt to diverse dehazing tasks\cite{b57}. This dynamic filtering enhances flexibility and improves performance across different scenarios.

\subsection{Residual Blocks}

Residual blocks (RB) are fundamental in modern CNNs, effectively addressing gradient issues in deep networks like ResNet\cite{b13}. Building on this, MemNet\cite{b29} introduced memory blocks (MB) that connect intermediate results across layers, while SRDenseNet\cite{b30} extended this by incorporating all preceding intermediate results before each convolution layer. RDN\cite{b31} further enhanced this design by employing 1×1 convolutions for feature fusion.

However, these models typically use convolution kernels of the same scale, which limits the diversity of receptive fields. LKD-Net\cite{b26} showed that dilated convolutions could achieve large receptive fields without increasing computational cost, inspiring the development of our residual multiscale block (RMB). The RMB employs convolution kernels with varying dilation rates to progressively fuse multiple receptive fields, improving the model's representation capability across different scales.

\subsection{Attention Mechanisms}

\begin{figure*}[t] 
  \centering
  \includegraphics[width=\textwidth]{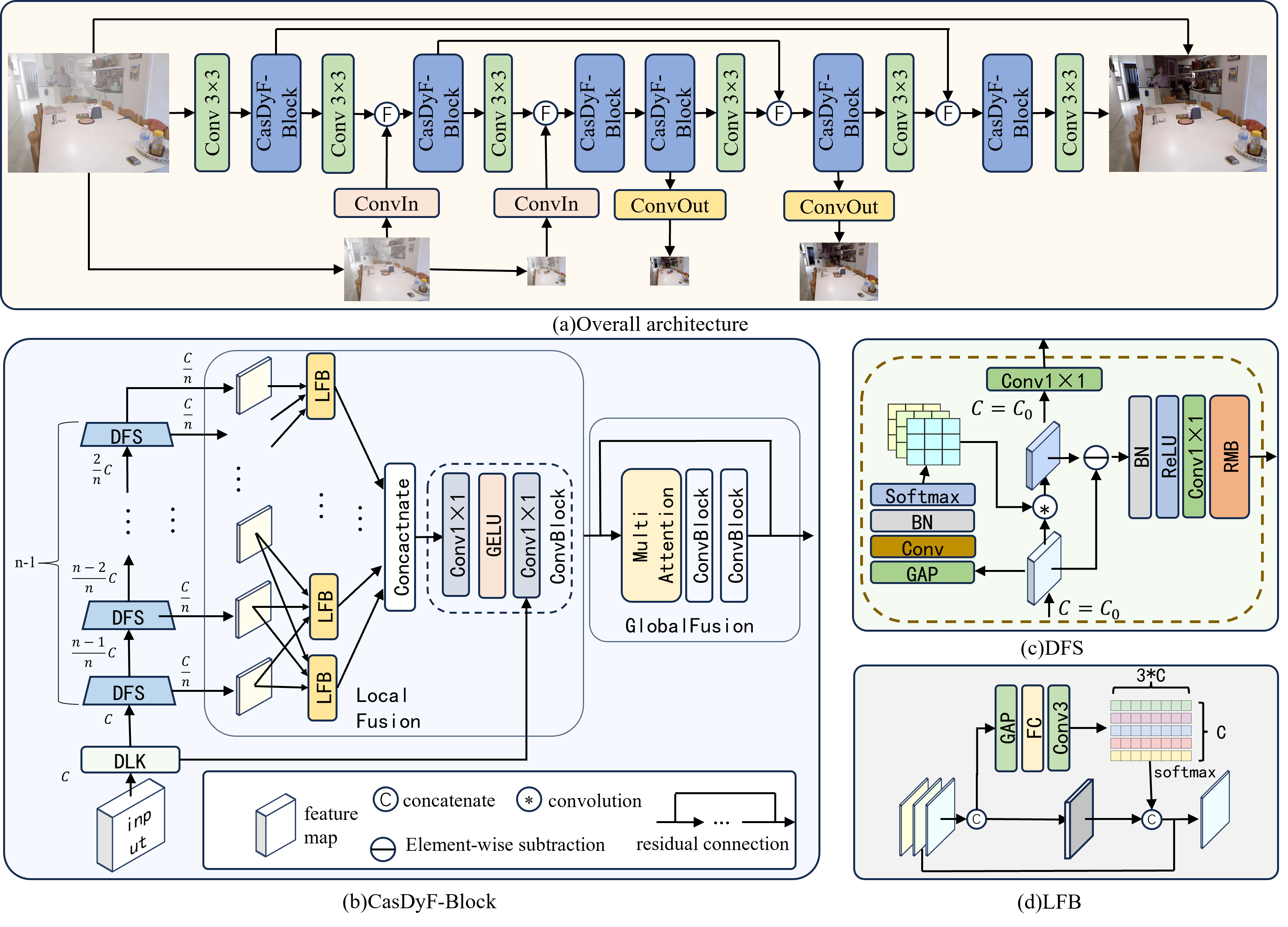} 
  \caption{The Proposed CasDyF-Net Network Architecture.(a) CasDyF-Net employs a popular U-shape structure, where the CasDyF-Block is our proposed Cascade Dynamic Filtering block.(b) The proposed CasDyF-Block consists of three processes: Dynamic Segmentation, Local Fusion, and Global Fusion. Dynamic Segmentation includes Dynamic Filtering and our proposed RMB (Residual Multiscale Convolution).(c) DFS (Dynamic Filtering and Segmentation) divides the feature maps into two parts using dynamic filtering.(d) The proposed Local Fusion Module utilizes dynamic 1 convolutions to fuse three adjacent feature branches into the current branch, with a residual connection added to the current branch..}
  \label{model}
\end{figure*}

Attention mechanisms are widely adopted in computer vision tasks to highlight important features and improve model performance. In image dehazing, various types of attention, including spatial\cite{b44} and channel attention\cite{b43}, have been used. More complex mechanisms, such as dual-domain attention\cite{b54}, have also been explored. Some models, like FFA-Net\cite{b22} and MixDehazeNet\cite{b27}, combine different attention types for better feature integration. Our model employs a progressive attention structure, combining our proposed local fusion block and existing mixed attention as global fusion. This structure integrates different feature branches more effectively and provides superior dehazing results by utilizing the strengths of both attention types.

\section{Proposed Method}
In this section, we will provide a detailed introduction to the proposed dehazing network. We will first present the overall structure of the CasDyF-Net, followed by an in-depth explanation of the implementation of each module. Finally, we will discuss the loss functions used in our approach.

\subsection{Overall Structure}

The architecture of CasDyF-Net, shown in Fig.~\ref{model}(a), uses a U-shaped structure with encoders and decoders based on the proposed CasDyF-Block. The CasDyF-Block creates multiple branches, incorporating modules like DFS, LFB, and RMB. DFS dynamically filters inputs to form branches, while RMB extracts features. The LFB preliminarily merges features from adjacent branches, followed by the global fusion module with mixed attention to further integrate features. 

The model takes a three-channel hazy image as input, shaped $H \times W \times 3$, where $H$ and $W$ are the image dimensions. Convolutional layers extract features and adjust channel numbers; for instance, the first convolutional layer increases the channel count to $C$, changing the feature map shape to $H \times W \times C$.

The network includes several skip connections. Information from the first two encoders is added as residuals to the features before the last two decoders. Additionally, following \cite{b57}, two low-resolution images are input to aid in learning low-level feature representations. After the last two decoders, two convolutional layers reconstruct low-resolution images, guiding the network to gradually restore a clear image.

\subsection{Dynamic Feature Segmentation (DFS)}
DFS dynamically separates fine-grained frequency components from feature maps, as illustrated in Fig.~\ref{model}(c). Dynamic filters, characterized by their frequency-selective properties \cite{b57}, decompose feature maps into two distinct frequency components. The dynamic filter system operates by generating convolutional kernels tailored to the input feature maps, which are subsequently convolved with the original feature maps. Given the input feature map $X_i$ at the $i$-th DFS level, the process of dynamic filtering is conducted as follows:

\begin{equation}
K_i=S\left(\mathit{BN}\left(W_i\left(\mathit{GAP}\left(X_i\right)\right)\right)\right)\text{, }i=1,2,...n-1\\
\end{equation}

\begin{equation}
Y_i=K_i\ast X_i\\
\end{equation}

\noindent where $GAP$, $W_i$, $BN$, and $S$ represent Global Average Pooling, convolutional layer parameters, Batch Normalization, and the Softmax function, respectively. $n-1$ is the number of cascaded DFS units. $K_i$ is the generated filter kernel, $Y_i$ is the filtered result, and $*$ denotes convolution. The obtained $Y_i$ is a separate feature branch. To refine it, we use a convolutional layer to reduce its channels, followed by RMB for feature extraction:

\begin{align}
F_i=\mathit{RMB}_i(W_i^{\mathit{out}}(\mathit{ReLU}(\mathit{BN}(Y_i)))) \text{,} i=1,..,n-1
\end{align}

\begin{equation}
F_n=X_n
\end{equation}

\noindent where ${RMB}_i$ represents our proposed residual multiscale block, and $W_i^{out}$ denotes the parameters of the 1×1 convolution used to reduce channels. At this stage, the feature branch is successfully isolated from $X_i$ and $F_i$.

Separating branch $Y_i$ from the sequential path $X_i$ reduces the information content in $X_i$. To utilize this information better, we do not directly use $X_i$ as input for the next DFS level. Instead, we use the complementary feature map $X_{i} - Y_i$ (as used in frequency selection\cite{b57}), then reduce its channels:

\begin{equation}
\begin{matrix}X_{i+1}=W_i^{\mathit{next}}\left(X_i-Y_i\right)
\end{matrix}
\end{equation}

\noindent where $W_i^{next}$ represents the convolutional layer parameters used to reduce channels. The entire feature segmentation process can be formulized as:

\begin{equation}
\left\{
\begin{aligned}
F_1^{\frac{C}{n}},\ X_2^{\frac{n-1}{n}C} &= \mathit{DF}S_1\left(X_1^C\right) \\
F_2^{\frac{C}{n}},\ X_3^{\frac{n-2}{n}C} &= \mathit{DF}S_2\left(X_2^{\frac{n-1}{n}C}\right) \\
&\vdots \ \\
F_{n-1}^{\frac{C}{n}},\ X_n^{\frac{C}{n}} &= \mathit{DF}S_{n-1}\left(X_{n-1}^{\frac{2}{n}C}\right) \\
F_n^{\frac{C}{n}} &= X_n^{\frac{C}{n}}
\end{aligned}
\right.
\end{equation}

\noindent where $F$ is the feature map separated in (3), and $X$ is the input feature map to the DFS. Their superscripts denote the number of channels in the feature maps.

\subsection{Residual Multiscale Block (RMB)}

After frequency segmentation, refining the segmented features is crucial. Large kernel convolutions are effective but lack flexibility and increase model size. Some approaches achieve similar effects by combining dilated convolutions. In view of that different convolutional scales focus on different frequency bands, we designed a multi-scale convolution module to refine selected frequency bands and introduced a 1×1 convolution module to merge information from different scales. The RMB includes three 3×3 convolutional kernels with different dilation rates, allowing for receptive fields ranging from fine to coarse. These kernels are sensitive to different frequency bands. The outputs of these convolutional kernels are combined through two 1×1 convolution layers.

\subsection{Progressive Attention Fusion}

To effectively fuse features from different branch, we designed a Progressive Feature Fusion module with two stages. In Local Fusion, each branch is fused with its adjacent branches. We dynamically generate separate parameters for each branch, enabling dynamic 1×1 convolutions, which enhance fusion flexibility. The process is as follows:

\begin{equation}
\begin{aligned}
F_i^l &= 
\begin{cases}
\mathit{LFB}_i\left(F_i,F_{i+1},F_{i+2}\right), & i=1 \\
\mathit{LFB}_i\left(F_{i-1},F_i,F_{i+1}\right), & i=2,3,\dots{},n-1 \\
\mathit{LFB}_i\left(F_{i-2},F_{i-1},F_i\right), & i=n
\end{cases}
\end{aligned}
\end{equation}

\noindent where $F_i^l$ is the result of local fusion, and ${LFB}_i$ represents the Local Fusion Block. The function of ${LFB}_i$ is to weightedly combine three groups of feature maps, achieving the initial fusion of adjacent branches.

The next stage is Global Fusion. To thoroughly fuse all branches, a globally adaptive attention module is required. Many approaches use mixed attention, combining various attention types either in series or parallel. We adopt the Enhanced Parallel Attention\cite{b27}, proven feasible in literature, combined with multiple 1×1 convolutions as our Global Attention Fusion module.

\begin{figure*}[t] 
  \centering
  \includegraphics[width=\textwidth]{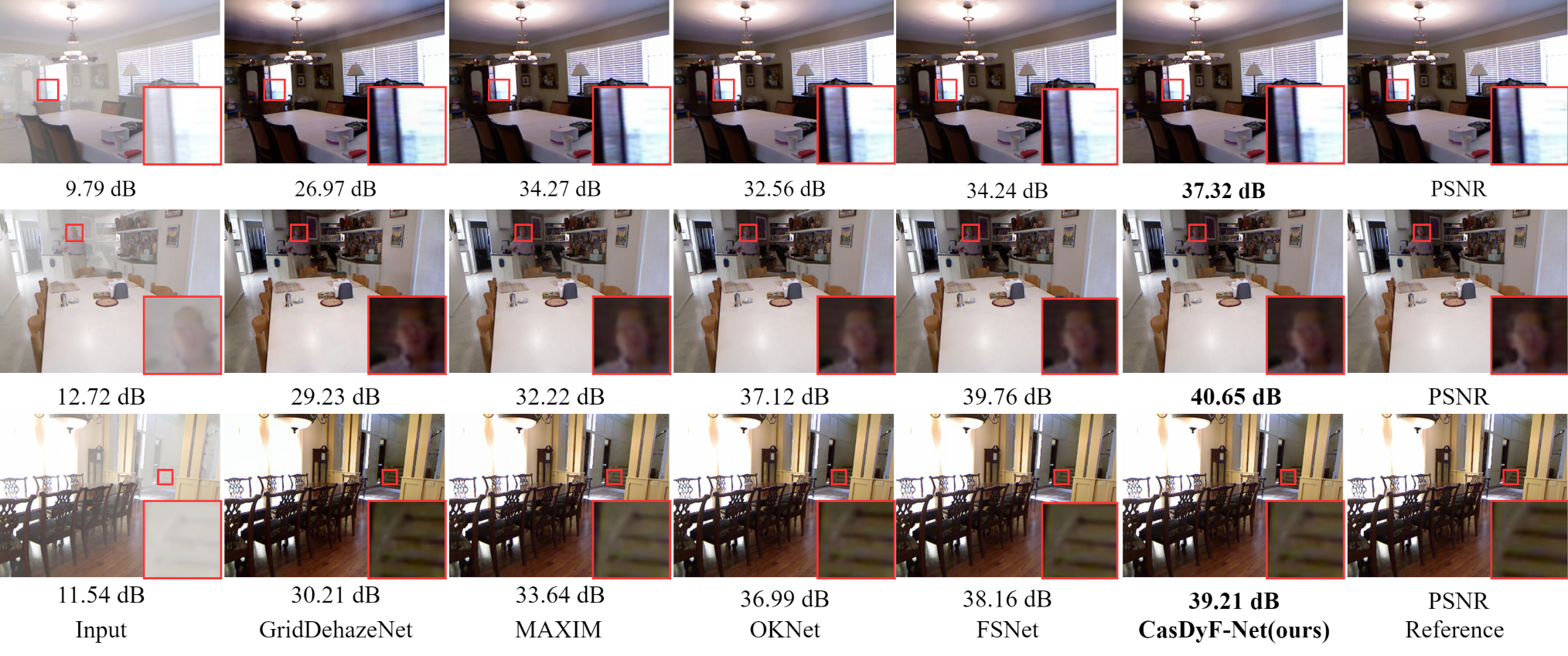} 
  \caption{Visual Comparison of Image Dehazing Effects on the SOTS-Indoor Dataset.}
  \label{its}
\end{figure*}

\begin{figure}[t]
\centerline{\includegraphics[width=0.5\textwidth]{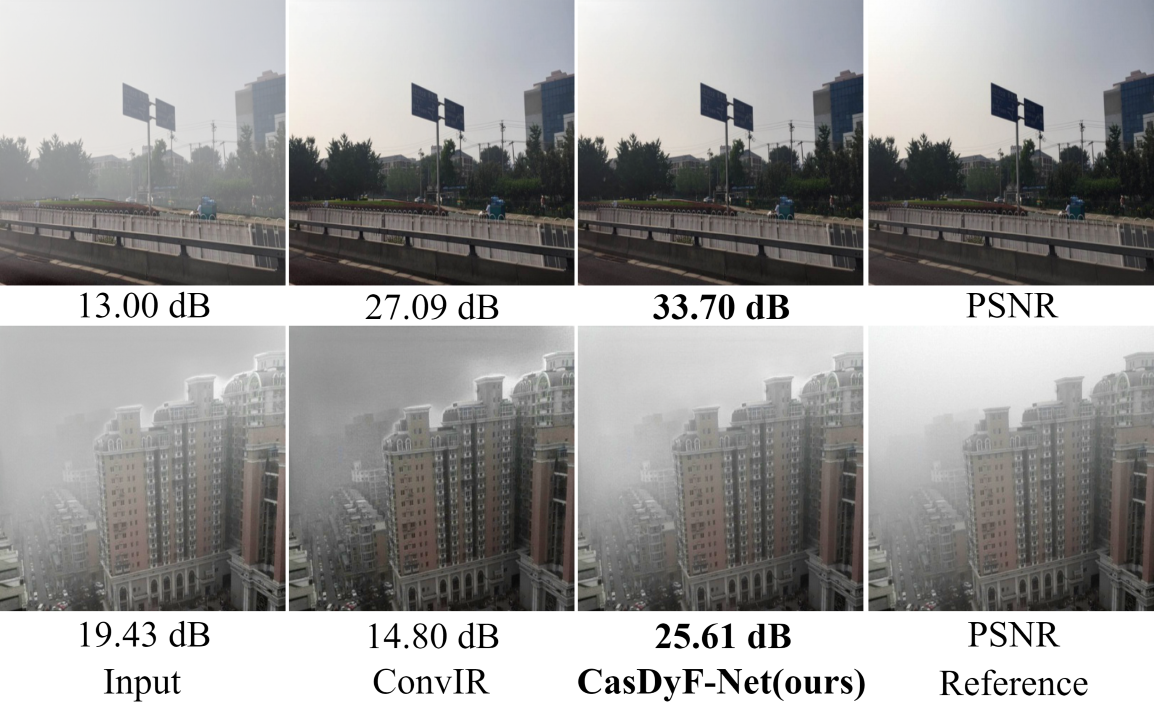}}
\caption{Visual Comparison on the Hzae4K Dataset.}
\label{haze4k}
\end{figure}

\begin{table}[t]
\caption{Comparison of different methods on RESIDE datasets}
\centering
\setlength{\tabcolsep}{4pt} 
\begin{tabular}{lcccccc}
\toprule
\multirow{2}{*}{Methods} &\multicolumn{2}{c}{SOTS-Indoor} & \multicolumn{2}{c}{SOTS-Outdoor} & Params & FLOPs \\ 
\cmidrule(lr){2-5}
&PSNR & SSIM & PSNR & SSIM & (M) & (G)\\
\midrule
GridDehazeNet\cite{b23} & 32.16 & 0.984 & 30.86 & 0.982 & 0.956 & 21.5 \\ 
FFA-Net\cite{b22} & 36.39 & 0.989 & 33.57 & 0.984 & 4.456 & 287.8 \\ 
MAXIM\cite{b62} & 38.11 & 0.991 & 34.19 & 0.985 & 14.1 & 216 \\ 
PMNet\cite{b24} & 38.41 & 0.990 & 34.74 & 0.985 & 18.9 & 81.13 \\ 
DehazeFormer-L\cite{b42} & 40.05 & 0.996 & - & - & 25.44 & 279.7 \\ 
OKNet\cite{b52} & 40.79 & 0.996 & 37.68 & 0.995 & 14.3 & 42 \\ 
DSANet\cite{b54} & 41.36 & \textbf{0.997} & 38.39 & 0.995 & 3.86 & 37.72 \\ 
FSNet\cite{b57} & 42.45 & \textbf{0.997} & \textbf{40.40} & \textbf{0.997} & 13.28 & 111 \\ 
MixDehazeNet-L\cite{b27} & 42.62 & \textbf{0.997} & 36.50 & 0.986 & 12.42 & 86.7 \\ 
MB-TaylorFormer-L\cite{b40} & 42.64 & 0.994 & 38.09 & 0.989 & 7.41 & 86.3 \\ 
ConvIR-B\cite{b51} & 42.72 & \textbf{0.997} & 39.42 & 0.996 & 8.63 & 71.22 \\ 
\midrule
CasDyF-Net (ours) & \textbf{43.21} & \textbf{0.997} & 38.86 & 0.995 & 6.23 & 40.55 \\ 
\bottomrule
\end{tabular}
\label{table:sots_comparison}
\end{table}

\begin{table}[t]
\caption{Comparison of different methods}
\centering
\begin{tabular}{lcccc}
\toprule
Methods & PSNR & SSIM & Params (M) & FLOPs (G) \\ 
\midrule
DehazeNet\cite{b63} & 19.12 & 0.84 & 0.01 & 0.58 \\ 
AOD-Net\cite{b64} & 17.15 & 0.83 & 0.002 & 0.12 \\ 
GridDehazeNet\cite{b23} & 23.29 & 0.93 & 0.956 & 21.5 \\ 
MSBDN\cite{b21} & 22.99 & 0.85 & 31.35 & 41.54 \\ 
FFA-Net\cite{b22} & 26.96 & 0.95 & 4.456 & 287.8 \\ 
DMT-Net\cite{b65} & 28.53 & 0.96 & - & - \\ 
PMNet\cite{b24} & 33.49 & 0.98 & 18.90 & 81.13 \\ 
FSNet\cite{b57} & 34.12 & \textbf{0.99} & 13.28 & 110.5 \\ 
ConvIR-L\cite{b51} & 34.50 & \textbf{0.99} & 14.83 & 129.34 \\ 
\midrule
CasDyF-Net (ours) & \textbf{35.73} & \textbf{0.99} & 6.23 & 40.55 \\ 
\bottomrule
\end{tabular}
\label{table:methods_comparison}
\end{table}

\subsection{loss Function}

The loss function we use accumulates the losses from three scales, with each scale’s loss containing both spatial domain loss and frequency domain loss. The specific formulation is as follows:

\begin{equation}
\begin{matrix}L=\sum _{s=1}^3\frac 1{E_s}\left(\left\|\widehat  X_s-X_s\right\|_1+\lambda \left\|\mathcal{F}\left(\widehat 
X_s\right)-\mathcal{F}\left(X_s\right)\right\|_1\right)\end{matrix}
\end{equation}

\noindent where $\widehat  X_s$ represents the output of the model at scale \textit{s}, and  $X_s$ is its corresponding ground truth.  $\mathcal{F}$ denotes the Fast Fourier Transform, and  $\left\|\cdot \right\|_1$ refers to the L1 loss function.  $\lambda $ is the coefficient for the frequency loss, set to 0.1 based on previous work\cite{b57}.

\section{Experiment}

\begin{figure*}[htbp] 
  \centering
  \includegraphics[width=\textwidth]{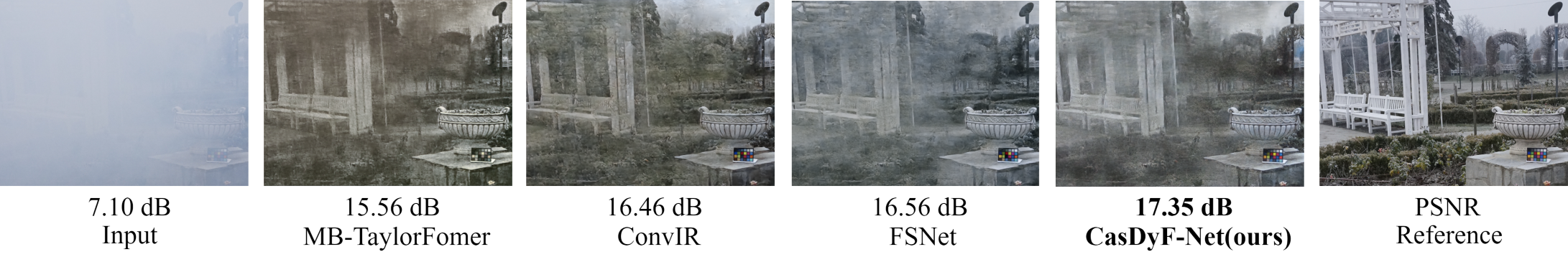} 
  \caption{Visual Comparison of Image Dehazing on the Dense-Haze Dataset.}
  \label{densevis}
\end{figure*}

\begin{figure}[t]
\centerline{\includegraphics[width=0.5\textwidth]{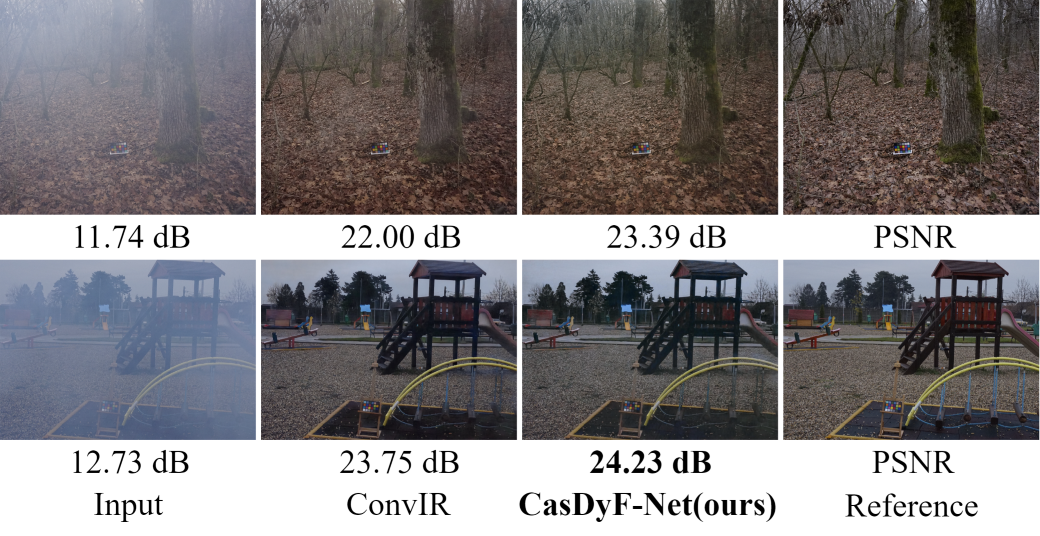}}
\caption{Visual Comparison of Image Dehazing on the O-HAZE Dataset.}
\label{ohaze}
\end{figure}

In this section, we will introduce and thoroughly analyze the experimental results of CasDyF-Net. First, we will present its performance across multiple datasets and compare it with other state-of-the-art models. Next, we will showcase the results of ablation studies to evaluate the effectiveness of each module. Finally, we will perform a qualitative analysis of the Residual Multiscale Block (RMB) in the frequency domain using the Fast Fourier Transform (FFT).
\subsection{Experimental Setup}
\paragraph{Details}

We trained our model on different datasets using a NVIDIA A800 GPU. For the ITS dataset \cite{b59} and Haze4K dataset \cite{b60}, the image patch size used for training was 256×256, with a batch size of 8, and the initial learning rate set to 4e-4. The training was conducted for 1000 epochs. On the DenseHaze and O-HAZE datasets, the image patch size was 600×800, with a batch size of 2, and the initial learning rate set to 2e-4, with training running for 5000 epochs. During training, we used a cosine annealing learning rate scheduler \cite{b58} to gradually reduce the learning rate to 1e-6.

\paragraph{Datasets}

On the ITS dataset, we used 13,990 hazy images for training and the 500 images from SOTS-indoor as the test set. For the OTS dataset, we used 313,740 images as the training set and 500 outdoor images from SOTS-outdoor as the test set. For the Haze4K dataset, we selected 3000 images as the    training set and 1000 images as the test set. The real-world datasets, Dense-Haze and O-HAZE, contain 55 and 45 paired images respectively, with the last 5 images from each dataset used as the test set, and the remaining images used for training.

\subsection{Experiments on Synthetic datasets}

We evaluated the performance of our model on the RESIDE dataset and two synthetic datasets, Haze4K, and compared it with the state-of-the-art models. The results are presented in Table~\ref{table:sots_comparison} and Table~\ref{table:methods_comparison}. The research findings show that our model outperforms the recent state-of-the-art models on both RESIDE-Indoor and Haze4K, achieving the best results in all aspects. Compared with Transformer-based methods such as MB-TaylorFormer-L and CNN-based methods such as ConvIR and FSNet, our model not only achieves better results but also significantly reduces the number of parameters and floating-point operations(FLOPs). In particular, compared with FSNet, which uses dynamic filtering, we achieve a 0.76dB ITS gain and a 1.61dB Haze4K gain with only 46.9\% of its FLOPs and 36.5\% of its parameters. Compared with the recent Transformer-based method MB-TaylorFormer-L, we achieve a 0.57dB ITS gain with only 84.1\% of its parameters and 47.0\% of its FLOPs.

Furthermore, we visually compared CasDyF-Net with other SOTA methods to show their haze removal effects (Fig.~\ref{its} and Fig.~\ref{haze4k}). Clearly, the images generated by our proposed model are closer to the reference images. 

\begin{table}[t]
\caption{Comparison of different methods on real datasets.}
\centering
\setlength{\tabcolsep}{6pt} 
\begin{tabular}{lcccc}
\toprule
\multirow{2}{*}{Methods} & \multicolumn{2}{c}{Dense-Haze} & \multicolumn{2}{c}{O-HAZE} \\ 
\cmidrule(lr){2-3} \cmidrule(lr){4-5}
&PSNR & SSIM & PSNR & SSIM \\ 
\midrule
GridDehazeNet\cite{b23}       & 13.31 & 0.368 & 18.92 & 0.672 \\ 
SGID-PFF\cite{b66}            & 12.49 & 0.517 & 20.96 & 0.741 \\ 
MSBDN\cite{b21}               & 15.13 & 0.555 & 24.36 & 0.749 \\ 
FFA-Net\cite{b22}             & 15.70 & 0.549 & 22.12 & 0.770 \\ 
DeHamer\cite{b37}             & 16.62 & 0.560 & 25.11 & 0.777 \\ 
PMNet\cite{b24}               & 16.79 & 0.510 & -     & -     \\ 
MB-TaylorFormer-L\cite{b40}   & 16.64 & 0.566 & 25.31 & 0.782 \\ 
ConvIR-B\cite{b51}            & 16.86 & 0.621 & 25.36 & 0.780 \\ 
\midrule
DFLS-Net (ours)     & \textbf{17.56} & \textbf{0.658} & \textbf{25.44} & \textbf{0.936} \\ 
\bottomrule
\end{tabular}
\label{table:dense}
\end{table}

\subsection{Experiments on Real datasets}

Additionally, we conducted further evaluation of CasDyF-Net on real-world datasets. The results demonstrate that our model exhibits leading performance on the real-world datasets compared to recently proposed techniques, achieving the best performance in both Dense-Haze and O-HAZE scenarios. Specifically, in the Dense-Haze scenario, our model outperforms other methods by 0.7dB; in the O-HAZE scenario, it leads by 0.08dB. It is worth noting that although our average PSNR advantage is relatively small in the O-HAZE scenario, the model demonstrates better stability, as evidenced by a significant advantage of 0.156 in structural similarity index (SSIM). This is because SSIM is less susceptible to individual sample effects.

In terms of visual effects, in the Dense-Haze scenario (Fig.~\ref{densevis}), both MB-TaylorFormer-L and ConvIR-B exhibit color difference issues, while FSNet shows unsatisfactory texture restoration in high-frequency areas such as forests. In contrast, our model balances overall color stability with high-frequency texture restoration capability. In the O-HAZE scenario (Fig.~\ref{ohaze}), our model significantly outperforms ConvIR in terms of texture restoration.

\subsection{Ablation Studies}
In this section, we first examine the effectiveness of each module to verify their contributions. Then, we explore several alternative solutions and conduct comparative analyses. Finally, we perform a qualitative study of some characteristics of the RMB. All experiments were conducted on the Haze4K dataset with 1000 training epochs.

\begin{table}[t]
\caption{Ablation Studies of Each Part.}
\centering
\setlength{\tabcolsep}{6pt} 
\begin{tabular}{lcccccc}
\toprule
Net & local & global & RMB & PSNR & Params(M) & FLOPs(G) \\ 
\midrule
(a) &   &   &   & 31.58 & 3.96 & 27.02 \\ 
(b) & $\checkmark$ &   &   & 33.80 & 4.83 & 28.63 \\ 
(c) &   & $\checkmark$ &   & 30.61 & 5.01 & 35.64 \\ 
(d) & $\checkmark$ & $\checkmark$ &   & 34.16 & 5.88 & 37.25 \\ 
(e) & $\checkmark$ & $\checkmark$ & $\checkmark$ & 35.73 & 6.23 & 40.55 \\ 
\bottomrule
\end{tabular}
\label{ablation}
\end{table}

\begin{table}[t]
\caption{Ablation Studies of RMB.}
\centering
\setlength{\tabcolsep}{8pt} 
\begin{tabular}{lcccc}
\toprule
Number of RMB & 0 & 1 & 2 & 3 \\ 
\midrule
PSNR (dB) & 34.16 & 35.39 & 35.73 & 35.59 \\ 
\bottomrule
\end{tabular}
\label{table:num_rmb}
\end{table}

\begin{table}[t]
\caption{Results of Alternatives to RMB.}
\centering
\setlength{\tabcolsep}{8pt} 
\begin{tabular}{lccccc}
\toprule
Method & None & RB & RDB & DLK7×7 & RMB \\ 
\midrule
PSNR (dB) & 34.16 & 34.39 & 34.43 & 35.13 & 35.39 \\ 
\bottomrule
\end{tabular}
\label{table:alt_rmb}
\end{table}

\begin{table}[t]
\caption{Comparison of Several Methods for Creating Branches.}
\centering
\setlength{\tabcolsep}{8pt} 
\begin{tabular}{lcccc}
\toprule
Method & Ours & Conv & Resolution & Split \\ 
\midrule
PSNR (dB)   & 35.39 & 34.31 & 34.15 & 33.39 \\ 
Params (M)  & 6.05  & 6.02  & 6.04  & 3.76  \\ 
FLOPs (G)   & 38.92 & 50.36 & 28.65 & 29.11 \\ 
\bottomrule
\end{tabular}
\label{table:alt_branch}
\end{table}

\paragraph{Effectiveness of Each Module}
As shown in Table~\ref{ablation}, the baseline model achieves a performance of 31.58 dB. When we introduced the proposed local attention module, the model’s performance improved by 2.22 dB, with only a 1 GFLOPs increase in computational cost. This indicates that the local fusion scheme is a successful design for CasDyF-Net. Additionally, we found that using global attention (i.e., EPA) alone actually resulted in lower performance compared to the baseline model. We hypothesize that this is due to the lack of residual connections, leading to excessive differences between branches, making it difficult for the global attention to effectively integrate all the information. However, when combined with the proposed local attention module, the model’s performance improved by 2.58 dB over the baseline model, demonstrating the success of the progressive fusion strategy. Furthermore, when we added the proposed RMB on this basis, the model’s performance was further enhanced by 1.57 dB, with only an additional 0.35M parameters and 3.3 GFLOPs, validating the effectiveness of the residual multiscale block.

\paragraph{Number of RMB}
To explore the impact of the number of RMBs on model performance, we adjusted the number of RMBs in each branch, as shown in Table~\ref{table:num_rmb}. Increasing the number of RMBs indeed improved the model’s PSNR, further validating the effectiveness of RMBs. However, we observed a performance drop when the number exceeded 2, likely due to overfitting. Therefore, our final model uses 2 RMBs.

\paragraph{Alternatives to RMB}
We also replaced the RMB with other advanced modules for comparison, to demonstrate the advantages of RMB, as shown in Table~\ref{table:alt_rmb}. The results indicate that the model with our multi-scale RMB outperformed models that did not use multi-scale schemes, such as RB\cite{b28} and RDB\cite{b31}. Compared to the novel dual-scale method LKD\cite{b26}, our RMB also achieved a significant performance improvement, indicating the effectiveness of the RMB design.

\paragraph{Alternatives to Cascaded Dynamic Filtering}
We also replaced the branch creation method with other common schemes and compared them with our proposed casca ded dynamic filtering method to test its advantages. The results are shown in Table~\ref{table:alt_branch}. First, to explore the lower bound of branch division schemes and establish a design baseline, we used the simplest channel splitting as the baseline. This method is used by some lightweight models, such as MFSN\cite{b49}. The baseline model achieved a performance of 33.39 dB.

Next, we replaced the dynamic filters with fixed filters, i.e., ordinary convolutional layers. We found that this approach increased the model’s computational overhead but resulted in poorer performance compared to dynamic filters. This is because fixed filters cannot adapt to the various distributions of hazy images, indirectly proving the advantage of dynamic filters.
We then applied the method used in ConvIR\cite{b51}, which creates a 4-branch network by progressively reducing the resolution. The experimental results show that this approach is also inferior to our proposed cascaded dynamic filtering method. While low-resolution images help the model understand low-frequency information, this scheme has inherent disadvantages for high-frequency information in the images.

\begin{figure*}[htbp] 
  \centering
  \includegraphics[width=\textwidth]{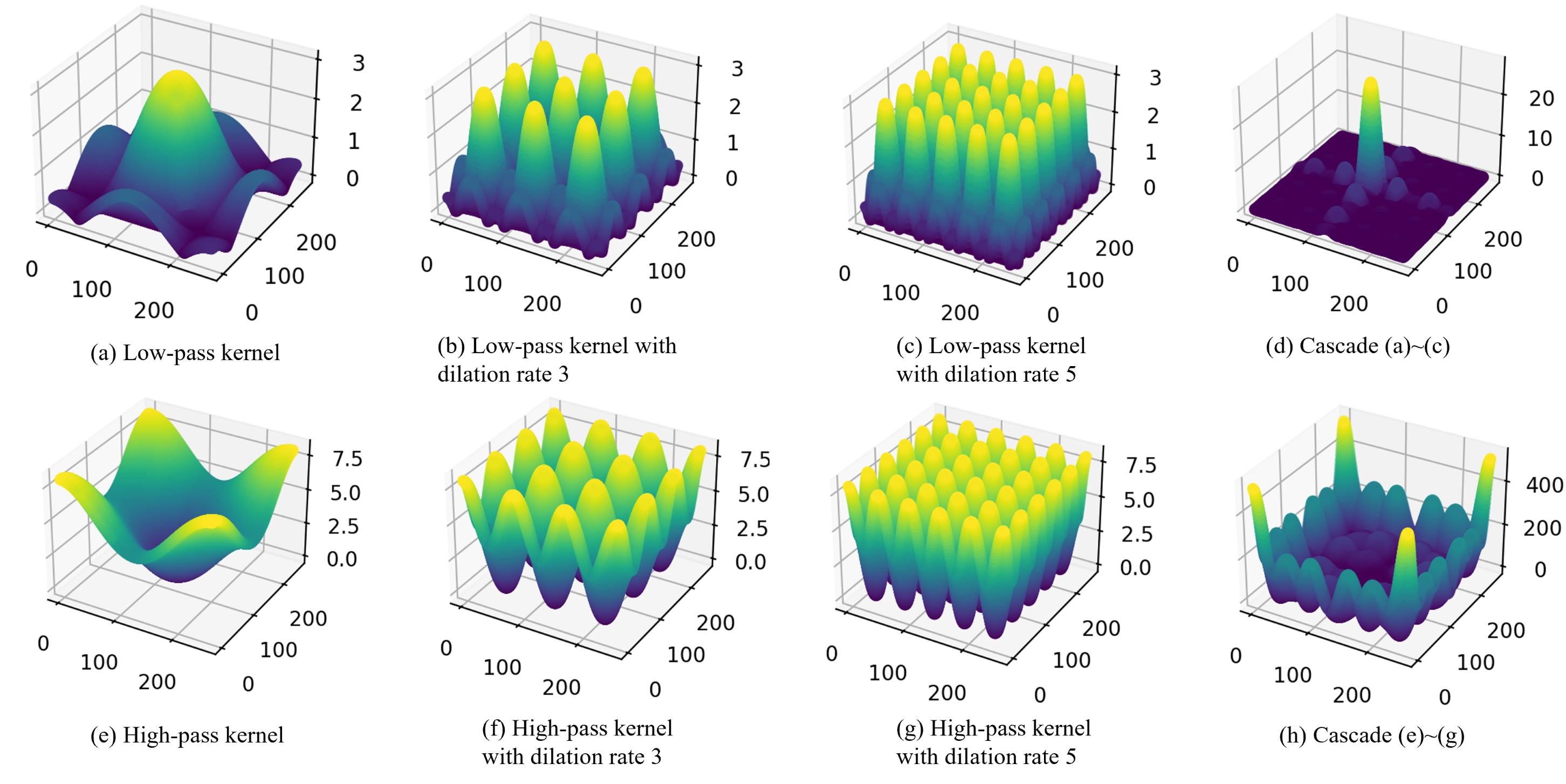} 
  \caption{Frequency response characteristics of several convolution kernels in RMB, where the center represents low frequency and the farther from the center, the higher the frequency. (a) Frequency spectrum of the average filter, which is a classic low-pass filter. (b) Frequency spectrum of the low-pass filter after a 3x dilation rate. (c) Frequency spectrum of the low-pass filter after a 5x dilation rate. (d) Frequency spectrum after cascading filters with different dilation rates. (e)$\sim$(h) Frequency spectra of the high-pass filter and its dilated versions.}
  \label{frequency}
\end{figure*}

\begin{figure}[t]
\centerline{\includegraphics[width=0.5\textwidth]{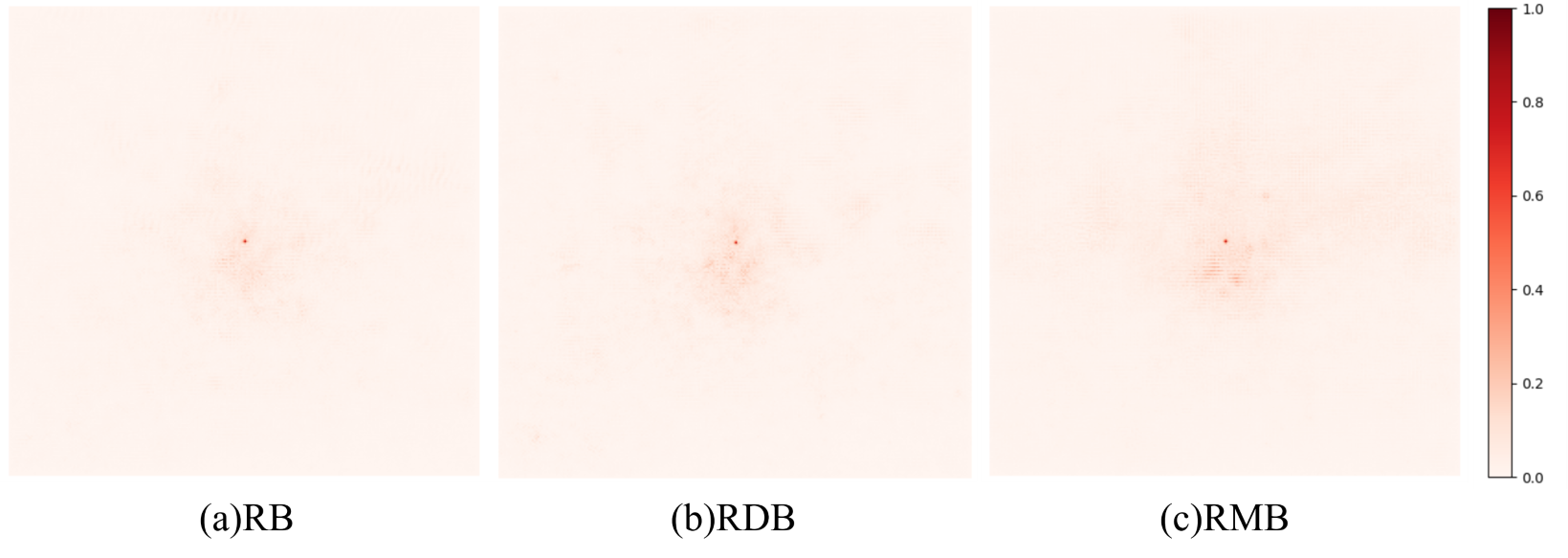}}
\caption{The Effective Receptive Fields (ERFs) of different version of models.}
\label{erf}
\end{figure}

\subsection{Qualitative analysis of RMB}
To explore the essence of dilated convolution and demonstrate the rationality of our proposed RMB, we designed a simple experiment to transform several different convolution kernels into the frequency domain to observe their frequency characteristics. We visualized the basic low-pass and high-pass filters (i.e., average filter and Laplacian edge detection filter), as shown in Fig.~\ref{frequency}. First, Fig.~\ref{frequency}(a) and Fig.~\ref{frequency}(e) show the original 3×3 filters, which, as expected, exhibit stronger passband characteristics in the low and high frequencies, respectively, as indicated by their peaks. When we increased the dilation rate, these patterns were compressed and repeated multiple times, resulting in more peaks. This indicates that compared to ordinary convolution, which can only focus on a single high or low frequency, dilated convolution can simultaneously focus on multiple frequency bands. However, there is also an obvious drawback: due to the repetition of the spectrum, the focus of dilated convolution on each frequency band is uniform. To leverage the advantages of filters with different dilation rates, we used them in series and then merged them in parallel. The resulting spectra, shown in Fig.~\ref{frequency}(d) and Fig.~\ref{frequency}(h), clearly demonstrate different levels of focus on multiple frequency bands, outperforming a single 3×3 convolution or dilated convolution.

Finally, to visually observe the influence of the dilated convolution on the RMB receptive field, we use the effective receptive field (ERF) theory proposed in \cite{b67} for visualization. As shown in Fig.~\ref{erf}, the network using our proposed RMB has a high sensitivity to a wider image area compared to the versions using RB\cite{b28} and RDB\cite{b31}.This means that our RMB has a larger effective receptive field.
\section{conclusion}

In conclusion, this paper presents CasDyF-Net, a novel image dehazing approach based on cascaded dynamic filters. Our method effectively addresses the limitations of traditional multi-branch networks by dynamically creating branches to capture diverse frequency features. The introduction of the Residual Multiscale Block (RMB) and a local fusion method based on dynamic convolution further enhances the model's ability to preserve texture details and integrate features across branches. Experimental results on multiple datasets demonstrate the superior performance of our model, achieving state-of-the-art results with reduced computational overhead. Our work contributes to the advancement of image dehazing technology, providing a more efficient and effective solution for restoring clarity to hazy images.

\end{document}